\documentclass[sigconf]{acmart}

\AtBeginDocument{%
  \providecommand\BibTeX{{%
    \normalfont B\kern-0.5em{\scshape i\kern-0.25em b}\kern-0.8em\TeX}}}

\usepackage{amsmath}
\usepackage{multirow}
\usepackage{subcaption}
\DeclareMathOperator*{\argmax}{arg\,max}

\setcopyright{iw3c2w3}

\begin{document}



\title{Recommending Themes for Ad Creative Design via Visual-Linguistic Representations}





\author{Yichao Zhou}
\affiliation{%
  \institution{University of California, Los Angeles
  }
}
\email{yz@cs.ucla.edu}

\author{Shaunak Mishra}
\affiliation{%
  \institution{Yahoo Research}
}
  \email{shaunakm@verizonmedia.com}

\author{Manisha Verma}
\affiliation{%
  \institution{Yahoo Research}
}
  \email{manishav@verizonmedia.com}
  
\author{Narayan Bhamidipati}
\affiliation{%
  \institution{Yahoo Research}
}
  \email{narayanb@verizonmedia.com}

\author{Wei Wang}
\affiliation{%
  \institution{University of California, Los Angeles}
}
  \email{weiwang@cs.ucla.edu}

\begin{abstract}
There is a perennial need in the online advertising industry to refresh ad creatives, \emph{i.e.}, images and text used for enticing online users towards a brand.
Such refreshes are required to reduce the likelihood of ad fatigue among online users, and to incorporate insights from other successful campaigns in related product categories.
Given a brand, to come up with themes for a new ad is a painstaking and time consuming process for creative strategists.
Strategists typically draw inspiration from the images and text used for past ad campaigns, as well as world knowledge on the brands.
To automatically infer ad themes via such multimodal sources of information in past ad campaigns,
we propose a theme (keyphrase) recommender system for ad creative strategists. The theme recommender is based on
aggregating results from a visual question answering (VQA) task, which ingests the following: (i) ad images, (ii) text associated with the ads as well as Wikipedia pages on the brands in the ads, and (iii) questions around the ad.
We leverage transformer based cross-modality encoders to train visual-linguistic representations for our VQA task. We study two formulations for the VQA task along the lines of classification and ranking; via experiments on a public dataset, we show that cross-modal representations lead to significantly better classification accuracy and ranking precision-recall metrics. 
Cross-modal representations show better performance compared to separate image and text representations. In addition, the use of multimodal information shows a significant lift over using only textual or visual information.

\end{abstract}


\copyrightyear{2020}
\acmYear{2020}
\acmConference[WWW '20]{Proceedings of The Web Conference 2020}{April 20--24, 2020}{Taipei, Taiwan}
\acmBooktitle{Proceedings of The Web Conference 2020 (WWW '20), April 20--24, 2020, Taipei, Taiwan}
\acmPrice{}
\acmDOI{10.1145/3366423.3380001}
\acmISBN{978-1-4503-7023-3/20/04}



\maketitle

\section{Introduction} \label{sec:introduction}
With the widespread usage of online advertising to promote brands (advertisers), there has been a steady need to innovate upon ad formats, and associated ad creatives \cite{wiki_banner_blindness}. The image and text comprising the ad creative can have a significant influence on online users, and their thoughtful design has been the focus of creative strategy teams assisting brands, advertising platforms, and third party marketing agencies.
\begin{figure}[t]
\centering
  \includegraphics[width=.9 \columnwidth]{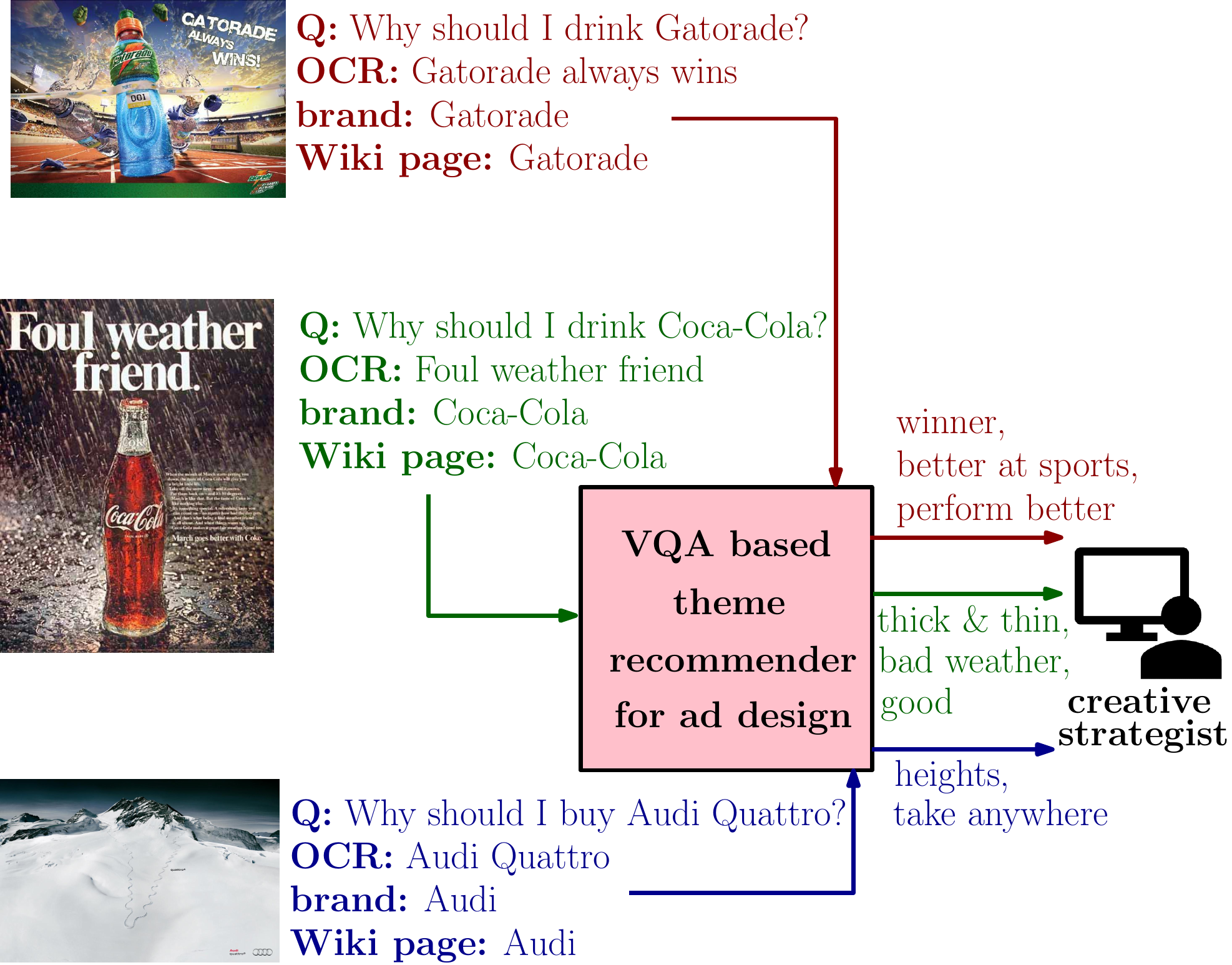}
  \caption{Ad (creative) theme recommender based on a VQA approach. The inputs are derived from past ad campaigns: (i) ad images, (ii) text in ad images (OCR), (iii) inferred brands, (iv) Wikipedia pages of inferred brands, and (v) questions around the ad. The recommended themes (keyphrases) are aggregated per brand (or product category), and can be used by a creative strategist to choose images (\emph{e.g.}, via querying stock image libraries) and text for new ads.}
  \label{fig:pull_figure}
  \vspace{-10pt}
\end{figure}
Numerous recent studies have indicated the emergence of a phenomenon called \textit{ad fatigue}, where online users get tired of repeatedly seeing the same ad each time they visit a particular website (\emph{e.g.}, their personalized news stream) \cite{ad_fatigue_schmidt,ad_fatigue_fb}. Such an effect is common even in native ad formats where the ad creatives are supposed to be in line with the content feed they appear in \cite{wiki_banner_blindness,ad_fatigue_fb}. In this context, frequently refreshing ad creatives is emerging as an effective way to reduce ad fatigue \cite{fb_business,marketing_land}.

From a creative strategist's view, coming up with new themes and translating them into ad images and text is a time taking task which inherently requires human creativity. Numerous online tools have emerged to help strategists in translating raw ideas (themes) into actual images and text, \emph{e.g.}, via querying stock image libraries \cite{shutterstock}, and by offering generic insights on the attributes of successful ad images and text \cite{taboola_trends}. In a similar spirit, there is room to further assist strategists by automatically recommending brand specific themes which can be used with downstream tools similar to the ones described above. In the absence of human creativity, inferring such brand specific themes using the multimodal (images and text) data associated with successful past ad campaigns (spanning multiple brands) is the focus of this paper.

A key enabler in pursuing the above data driven approach for inferring themes is that of a dataset of ad creatives spanning multiple advertisers. Such a dataset \cite{creative_dataset} spanning $64,000$ ad images was recently introduced in \cite{cvpr_kovashka}, and also used in the followup work \cite{kovashka_eccv2018}. The collective focus in the above works \cite{cvpr_kovashka,kovashka_eccv2018} was on understanding ad creatives in terms of sentiment, symbolic references and VQA. In particular, no connection was made with the brands inferred in creatives, and the associated world knowledge on the inferred brands. As the first work in connecting the above dataset \cite{creative_dataset} with brands, \cite{self_recsys2019} formulated a \textit{keyword} ranking problem for a brand (represented via its Wikipedia page), and such keywords could be subsequently used as themes for ad creative design. However, the ad images were not used in \cite{self_recsys2019}, and recommended themes
were restricted to single words (keywords) as opposed to longer keyphrases which could be more relevant. For instance, in Figure~\ref{fig:pull_figure}, the phrase \textit{take anywhere} has much more relevance for Audi than the constituent words in isolation.

In this paper, we primarily focus on addressing both the above mentioned shortcomings by
(i) ingesting ad images as well as textual information, \emph{i.e.},
Wikipedia pages of the brands and text in ad images (OCR), and (ii) we consider keyphrases (themes) as opposed to keywords. Due to the multimodal nature of our setup, we propose a VQA formulation as exemplified in Figure~\ref{fig:pull_figure}, where the questions are around the advertised product (as in \cite{kovashka_eccv2018,cvpr_kovashka}) and the answers are in the form of keyphrases (derived from answers in \cite{creative_dataset}).
Brand specific keyphrase recommendations can be subsequently collected from the predicted outputs of brand-related VQA instances. Compared to prior VQA works involving questions around an image, the difference in our setup lies in the use of Wikipedia pages for brands, and OCR features; both of these inputs are considered to assist the task of recommending ad themes.
In summary, our main contributions
can be listed as follows:
\begin{enumerate}
     \item we study two formulations for VQA based ad theme recommendation (classification and ranking) while using multimodal sources of information (ad image, OCR, and Wikipedia),
     \item we show the efficacy of transformer based visual-linguistic representations for our task, with significant performance lifts versus using separate visual and text representations,
     \item we show that using multimodal information (images and text) for our task is significantly better than using only visual or textual information, and
    \item we report selected ad insights from the public dataset \cite{creative_dataset}.
\end{enumerate}

The remainder of the paper is organized as follows. Section~\ref{sec:related} covers related work, and Section~\ref{sec:method} describes the proposed method. The data sources, results, and insights are then described in Section~\ref{sec:results},
and we conclude the paper with Section~\ref{sec:conclusion}.

\section{Related Work} \label{sec:related}
In this section, we cover related work on online advertising, understanding ad creatives, and visual-linguistic representations.
\vspace{-5pt}
\subsection{Online advertising}
Brands typically run online advertising campaigns in partnerships with publishers (\emph{i.e.}, websites where ads are shown), or advertising platforms~\cite{Google_FTRL,gemx_kdd} catering to multiple publishers. Such an ad campaign may be associated with one or more ad creatives to target relevant online users. 
Once deployed, the effectiveness of targeting and ad creatives is jointly gauged via metrics like click-through-rate (CTR), and conversion rate (CVR)~\cite{mappi_CIKM}. To separate out the effect of targeting from creatives, advertisers typically try out different creatives for the same targeting segments, and efficiently explore \cite{explore_exploit_li} which ones have better performance. In this paper, we focus on quickly creating a pool of ad creatives for a brand (via recommended themes learnt from past ad campaigns), which can then be tested online with targeting segments.

\vspace{-5pt}
\subsection{Automatic understanding of ad creatives}
The \textit{creatives dataset}~\cite{creative_dataset} is one of the key enablers of the proposed recommender system. This dataset was introduced in~\cite{cvpr_kovashka}, where the authors focused on automatically understanding the content in ad images and videos from a computer vision perspective. The dataset has ad creatives with annotations including topic (category), questions and answers (\emph{e.g.}, reasoning behind the ad, expected user response due the ad).  In a followup work~\cite{kovashka_eccv2018}, the focus was on understanding symbolism in ads (via object recognition and image captioning) to match human-generated statements describing actions suggested in the ad. Understanding ad creatives from a brand's perspective was missing in both \cite{kovashka_eccv2018, cvpr_kovashka}, and \cite{self_recsys2019} was the first to study the problem of recommending keywords for guiding a brand's
creative design. However, \cite{self_recsys2019} was limited to only text inputs for a brand (\emph{e.g.}, the brand's Wikipedia page), and the recommendation was limited to single words (keywords).
In this paper, we extend the setup in \cite{self_recsys2019} in a non-trivial manner by including multimodal information from past ad campaigns, \emph{e.g.}, images, text in the image (OCR), and Wikipedia pages of associated brands. We also extend the recommendations from single words to longer keyphrases.
\vspace{-5pt}
\subsection{Visual-linguistic representations and VQA}
With an increasing interest in joint vision-language tasks like visual question answering (VQA)~\cite{vqa_iccv2015}, and image captioning~\cite{sharma2018conceptual}, there has been lot of recent work on visual-linguistic representations which are key enablers in the above mentioned tasks. In particular, there has been a surge of proposed methods using transformers~\cite{devlin2018bert}, and we cover some of them below.

In LXMERT~\cite{lxmert}, the authors proposed a transformer based model that encodes different relationships between text and visual inputs trained using five different pre-training tasks. More specifically, they use encoders that model text, objects in images and relationship between text and images using (image,sentence) pairs as training data. They evaluate the model on two VQA datasets. More recently ViLBERT~\cite{vilbert} was proposed, where BERT~\cite{devlin2018bert} architecture was extended to generate multimodal embeddings by processing both visual and textual inputs in separate streams which interact through co-attentional transformer layers. The co-attentional transformer layers ensure that the model learns to embed the interactions between both modalities. Other similar works include VisualBERT~\cite{li2019visualbert}, VLBERT~\cite{su2019vl}, and Unicoder-VL~\cite{li2019unicoder}. 

\begin{figure*}[ht]
\centering
  \includegraphics[width=1.3 \columnwidth]{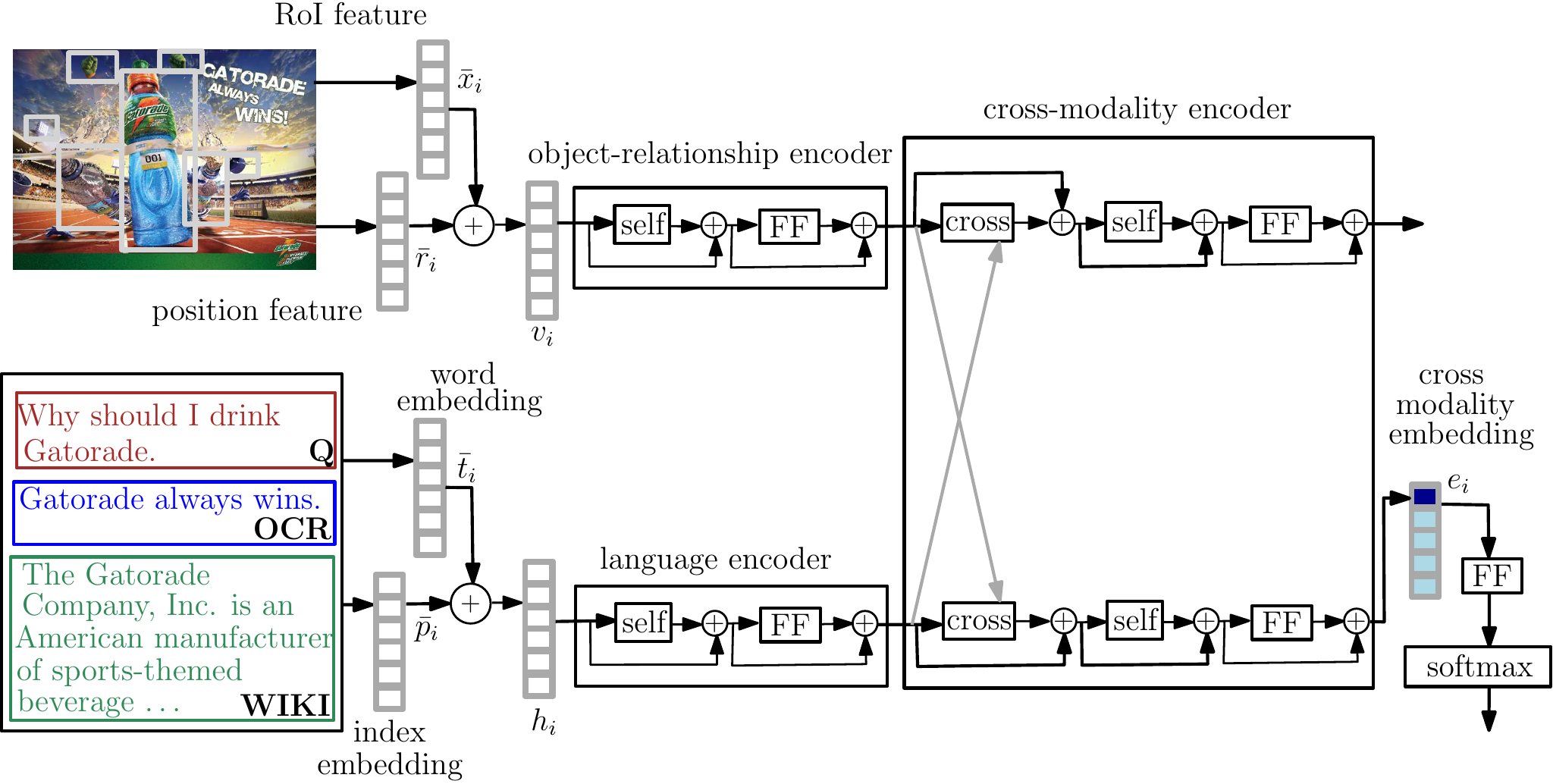}
  \caption{Cross modality 
  encoder architecture, and subsequent feed forward (FF) network with softmax layer for the classification objective.}
  \label{fig:lxmert_model}
\end{figure*}

In this paper, our goal is to focus on leveraging
visual-linguistic representations 
to solve an ads specific VQA task formulated to infer brand specific ad creative themes. In addition, VQA tasks on ad creatives tend to relatively challenging (\emph{e.g.}, compared to image captioning)
due to the subjective nature and hidden symbolism frequently found in ads~\cite{kovashka_eccv2018}.
Another difference between our work and existing VQA literature is that our task is not limited to 
understanding the objects in the image but also the emotions the ad creative would evoke in the reader. Our primary task is to predict different themes and sentiments that an ad creative image can invoke in its reader, and use such brand specific understanding to help creative strategists in developing new ad creatives.

\section{Method} \label{sec:method}
We first describe our formulation of ad creative theme recommendation as a classification problem in Section~\ref{sec:classification_formulation}. This is followed by subsections on text and image representation, cross modal encoder, and optimization
(Figure~\ref{fig:lxmert_model} gives an overview).
Finally, in Section~\ref{sec:ranking_formulation} we cover an alternative ranking formulation for recommendation.

\subsection{Theme recommendation: classification formulation} \label{sec:classification_formulation}
In our setup, we are given an ad image $X_{i}$ (indexed by $i$),
and associated text denoted by $S_i$. Text $S_i$ is sourced from: (i) text in ad image (OCR), (ii) questions around the ad, and (iii) Wikipedia page of the brand in the ad.
Given $X_i$, we represent the image as a sequence of objects $x_i = \{x_{i,1}, x_{i,2}, ..., x_{i,n}\}$ together with their corresponding regions in the image $r_i = \{r_{i,1}, r_{i,2}, ..., r_{i,n}\}$
(details in Section~\ref{sec:image_text_embeddings}).
The sentence $S_i$ is represented as a sequence of words $w_i = \{w_{i,1}, w_{i,2}, ..., w_{i,m}\}$. Given the three sequences $x_i,r_i,w_i$, the objective is to recommend a keyphrase $\hat{k}_i \in \mathcal{K}$, where $\mathcal{K}$ is a pre-determined vocabulary of keyphrases. In other words, for $k \in \mathcal{K}$, the goal is to estimate the probability $\mathbb{P}(k | x_i, r_i, w_i)$, and then the top keyphrase $\hat{k}_i$ for instance $i$ can be selected as:
\begin{align}
    \hat{k}_i  = \argmax_{k \in \mathcal{K}} \; \mathbb{P} (k | x_i, r_i, w_i).
\end{align}
The above classification formulation is similar to that for VQA in \cite{kovashka_eccv2018}; the difference is in the multimodal features explained below.
\subsection{Text and image embeddings} \label{sec:image_text_embeddings}
\noindent \textbf{Text embedding.}
We first use WordPiece Tokenizer~\cite{wu2016google} to convert a sentence $w_i$ into a sequence of tokens $t_i = \{t_{i,1}, t_{i,2}, ..., t_{i,l}\}$. Then, the tokens are projected to vectors in the embedding layer leading to $\bar{t}_i$ (as shown in \eqref{eq:t_emb}). Their corresponding positions $p_i$ are also projected to vectors leading to $\bar{p}_i$ (as shown in \eqref{eq:p_emb}). Then, $\bar{t}_i$ and $\bar{p}_i$ are added to form $h_i$ as shown in \eqref{eq:h} below:
\begin{align}
        \bar{t}_i &= E_t * t_i, \label{eq:t_emb}
        \\
        \bar{p}_i &= E_p * p_i, \label{eq:p_emb} \\
        h_i &= 0.5 * (\bar{t}_i + \bar{p}_i), \label{eq:h}
    \end{align}
where $E_t\in \mathbb{R}^{|V_t| \times D_t}$ and $E_p\in \mathbb{R}^{|V_p| \times D_p}$ are the embedding matrices. $|V_t|$ and $|V_p|$ are the vocabulary size of tokens, and token positions. $D_t$ and $D_p$ are the dimensions of token and position embeddings.

\noindent \textbf{Image embedding.}
We use bounding boxes and their region-of-interest (RoI) features to represent an image. 
Same as \cite{vilbert,lxmert}, we leverage Faster R-CNN~\cite{ren2015faster} to generate the bounding boxes and RoI features. Faster R-CNN is an object detection tool which identifies instances of objects belonging to certain classes, and then localizes them with bounding boxes. Though image regions lack a natural ordering compared to token sequences, the spatial locations can be encoded (\emph{e.g.}, as demonstrated in \cite{lxmert}). The image embedding layer takes in the RoI features $x_i$ and spatial features $r_i$ and outputs a position-aware image embedding $v_i$ as shown below:
\begin{align}
        \bar{x}_i &=  W_x * x_i + b_x,  \nonumber \\
        \bar{r}_i &=  W_r * r_i + b_r, \nonumber \\ 
        v_i &= 0.5 * (\bar{x}_i + \bar{r}_i),
\end{align}
where $W_x$ and $W_r$ are weights, and $b_x$ and $b_r$ are biases.

\subsection{Transformer-based cross-modality encoder} \label{sec:cross_modal_encoder}
We apply a transformer-based cross-modality encoder to learn a joint embedding from both visual and textual features. Here, without loss of generality, we follow the LXMERT architecture from \cite{lxmert} to encode the cross-modal features. As shown in Figure~\ref{fig:lxmert_model}, the token embedding $h_i$ is first fed into a language encoder while the image embedding $v_i$ goes through an object-relationship encoder. The cross-modality encoder contains two unidirectional cross-attention sub-layers which attend the visual and textual embeddings to each other. We use the cross-attention sub-layers to align the entities from two modalities and to learn a joint embedding $e_i$ of dimension $D_e$. We follow \cite{devlin2018bert} to add a special token [CLS] to the front of the token sequence. The embedding vector learned for this special token is regarded as the cross-modal embedding\footnote{Recently proposed ViLBERT \cite{vilbert}, and VisualBERT \cite{li2019visualbert} can serve as alternatives.}. In terms of query ($Q$), key ($K$), and value ($V$), the visual ${cross\text{-}attention}(Q, K, V) = \text{softmax}(\frac{QK^T}{\sqrt{d}})V$
where $Q$, $K$, and $V$ are linguistic features, visual features, and visual features, respectively; $d$ represents the dimension of linguistic features \cite{lxmert}. Textual cross-attention is similar with visual and linguistic features swapped.

\subsection{Learning and optimization.} \label{sec:optimization}
Based on the joint embedding for each image and sentence pair, the keyphrase recommendation task can now be tackled with a fully connected layer.
Given the cross-modal embedding, the probability distribution over all the candidate keyphrases is calculated by a fully-connected layer and the softmax function as shown below:
\begin{align}
    \hat{\mathbb{P}}(k|x_i, r_i , w_i) = \text{softmax}( W_f \cdot e_i + b_f)
\end{align}
where $W_f$ and $b_f$ are the weight and bias of a fully connected layer, and $e_i$ is the cross modal embedding. 

\subsection{Theme recommendation: ranking formulation} \label{sec:ranking_formulation}
\begin{figure}[!htb]
\vspace{-10pt}
\centering
  \includegraphics[width=0.55\columnwidth]{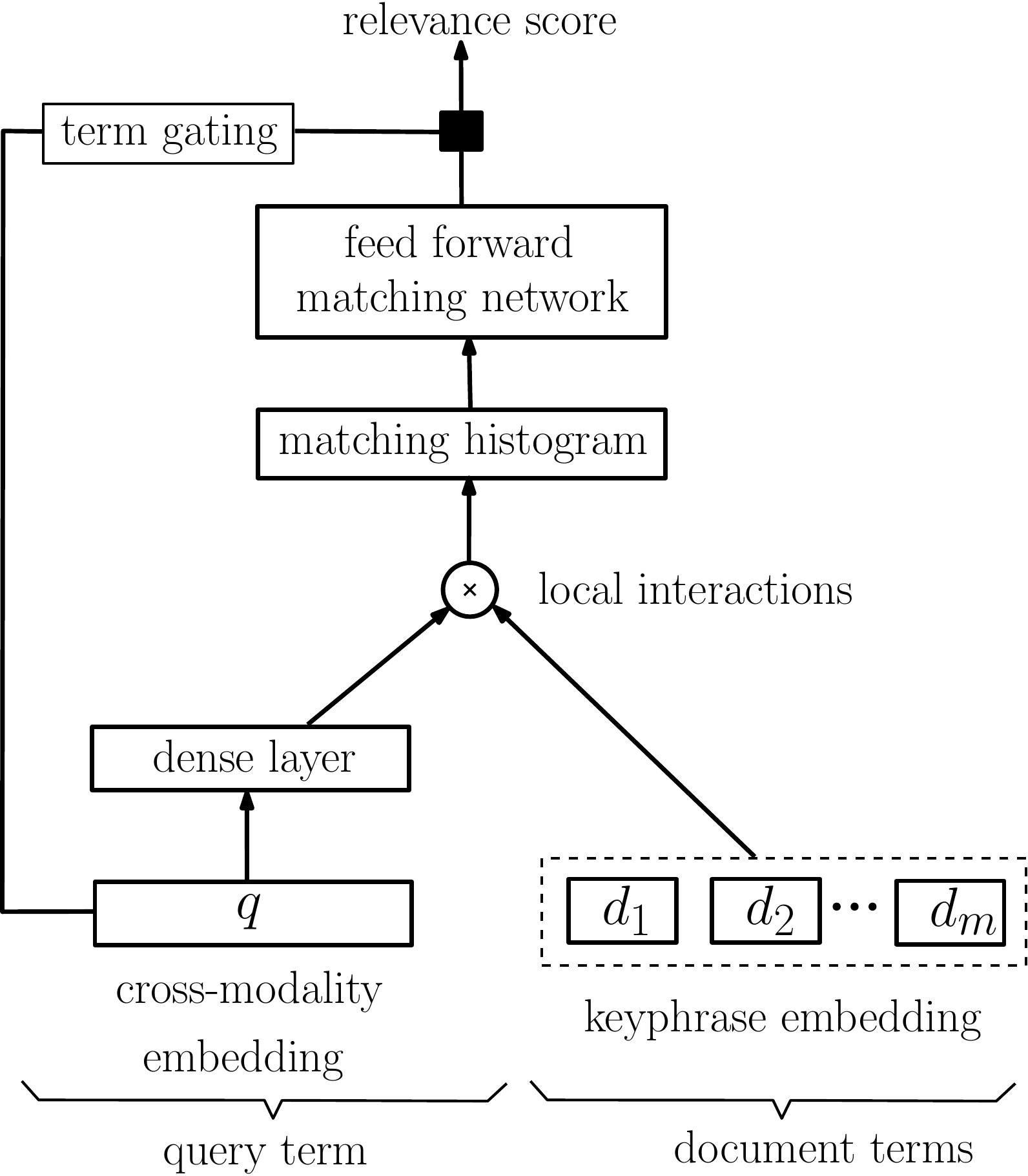}
  \caption{DRMM for the keyphrase ranking objective.}
  \label{fig:drmm}
  \vspace{-5 pt}
\end{figure}

We also consider solving the theme recommendation problem via a ranking model, where the model outputs a  list of keyphrases in decreasing order of relevance for a given \emph{(image, sentence)} pair, \emph{i.e.}, $(X_i, S_i)$. We use the state-of-the-art pairwise deep relevance matching model (DRMM)~\cite{drmm} whose architecture for our theme recommendation setup is shown in  Figure~\ref{fig:drmm}. It is worth noting that our pairwise ranking formulation can be changed to accommodate other multi-objective or list-based loss-functions. We chose the DRMM model since it is not restricted by the length of input, as most ranking models are, but relies on capturing local interactions between query and document terms with fixed length matching histograms. 
Given an \emph{(image, sentence, phrase)} combination, the model selects top k interactions between cross-modal embedding and the phrase embedding. These interactions are passed through a multilayer perceptron (MLP), and the overall score is aggregated with a query term gate which is a softmax function over all terms in that query.

The ranking segment of the model takes two inputs: (i) cross-modal embedding for $(X_i, S_i)$ (as explained in Section~\ref{sec:cross_modal_encoder}), and (ii) the phrase embedding. It then learns to predict the relevance of the given phrase with respect to the query \emph{(image, sentence)} pair.
Given that our input documents (\emph{i.e.}, keyphrases) are short,
we select top $\theta$ interactions in matching histogram between the cross-modal embedding, and the keyphrase embedding.
Mathematically, we denote the \emph{(image, sentence)} pair by just the \emph{imgq} below. Given a triple ($imgq$, $p^+$, $p^-$) where $p^+$ is ranked higher than $p^-$ with respect to image-question, the loss function is defined as:
\begin{align}
\mathcal{L} (imgq,p^+,p^-;\theta)=max(0, 1-s(imgq,p^+) + s(imgq,p^-)) ,
\end{align}
where $s(imgq, p)$ denotes the predicted matching score for phrase $p$, and the query image-question pair.  

\section{Experiments} \label{sec:results}
In this section, we go over the public dataset used in our experiments, classification and ranking results, and inferred insights.

\subsection{Dataset}
We rely on a publicly available data set \cite{cvpr_kovashka, creative_dataset} that consists of $64,000$ advertisement creatives, spanning $700$ brands across $39$ categories, among which 80\% is training set and 20\% is test set. We select 10\% data from the training set for validation. 
Crowdsourcing was used to gather following labels for each creative: (i) topics ($39$ types), (ii) questions and answers as reasons for buying from the brand depicted in the creative ($\sim$3 per creative). In addition to the existing annotations, we add the following annotations: (i) brand present in a creative, (ii) Wikipedia page relevant to the brand-category pair in a creative, and (iii) the set of target themes (keyphrases) associated with each image. In particular, for (i) and (ii) we follow the method in \cite{self_recsys2019}, and for (iii) the keyphrases (labels) were extracted from the answers using the position-rank method~\cite{pke,position-rank} for each image. The number of keyphrases was limited to at most $5$ (based on the top keyphrase scores returned by position-rank). We define a score for each keyphrase. All five keyphrases have scores of $1.0, 0.9, 0.8, 0.7,$ and $0.6$ in order \footnote{The annotated ads dataset can be found at \url{https://github.com/joey1993/ad-themes}.}.

\begin{figure}[ht]
\centering
    \begin{subfigure}[b]{0.48\linewidth}
        \includegraphics[width=1\linewidth]{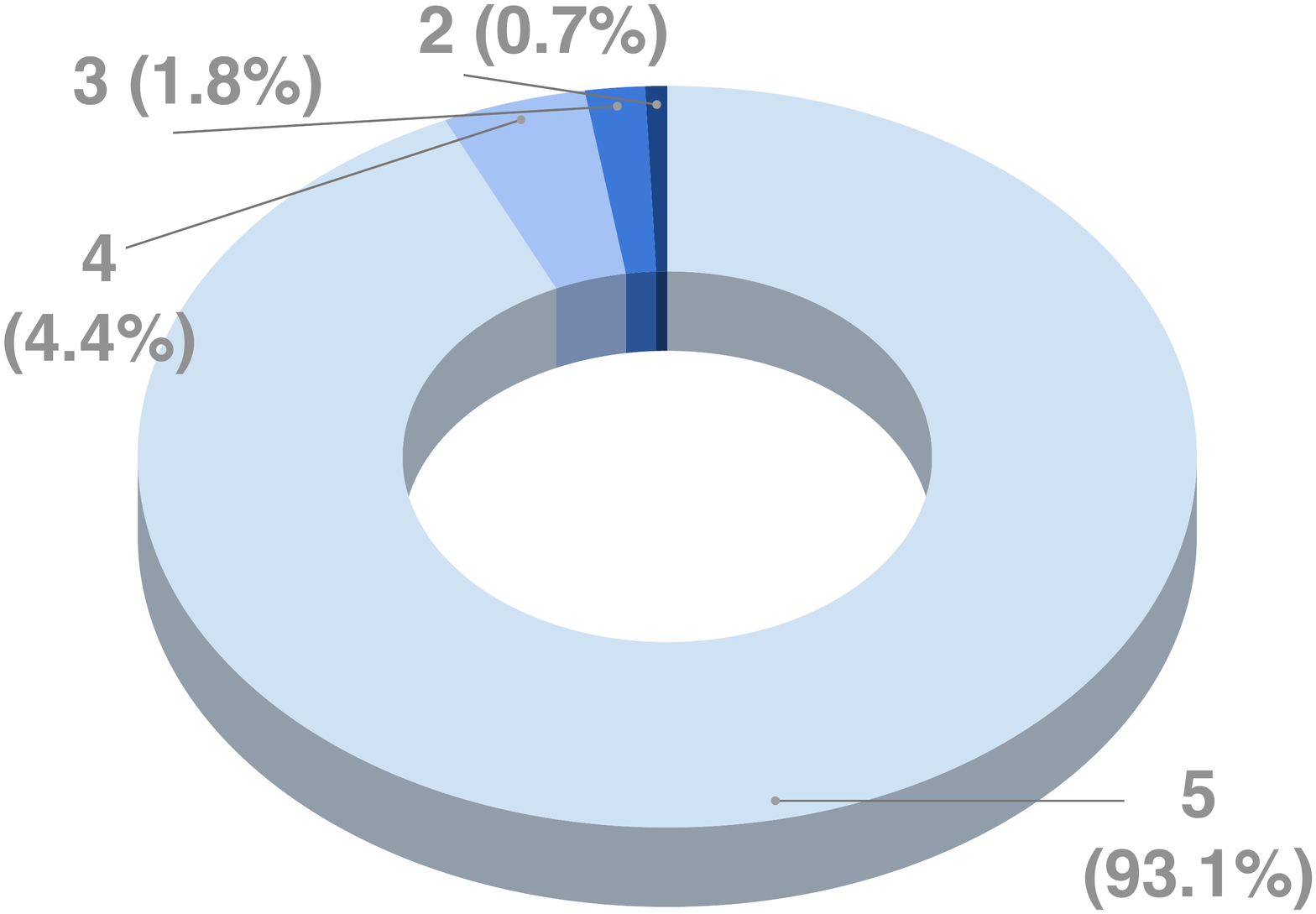}
        \caption{\centering \#keyphrases per instance}
    \end{subfigure}
    \begin{subfigure}[b]{0.48\linewidth}
        \includegraphics[width=1\linewidth]{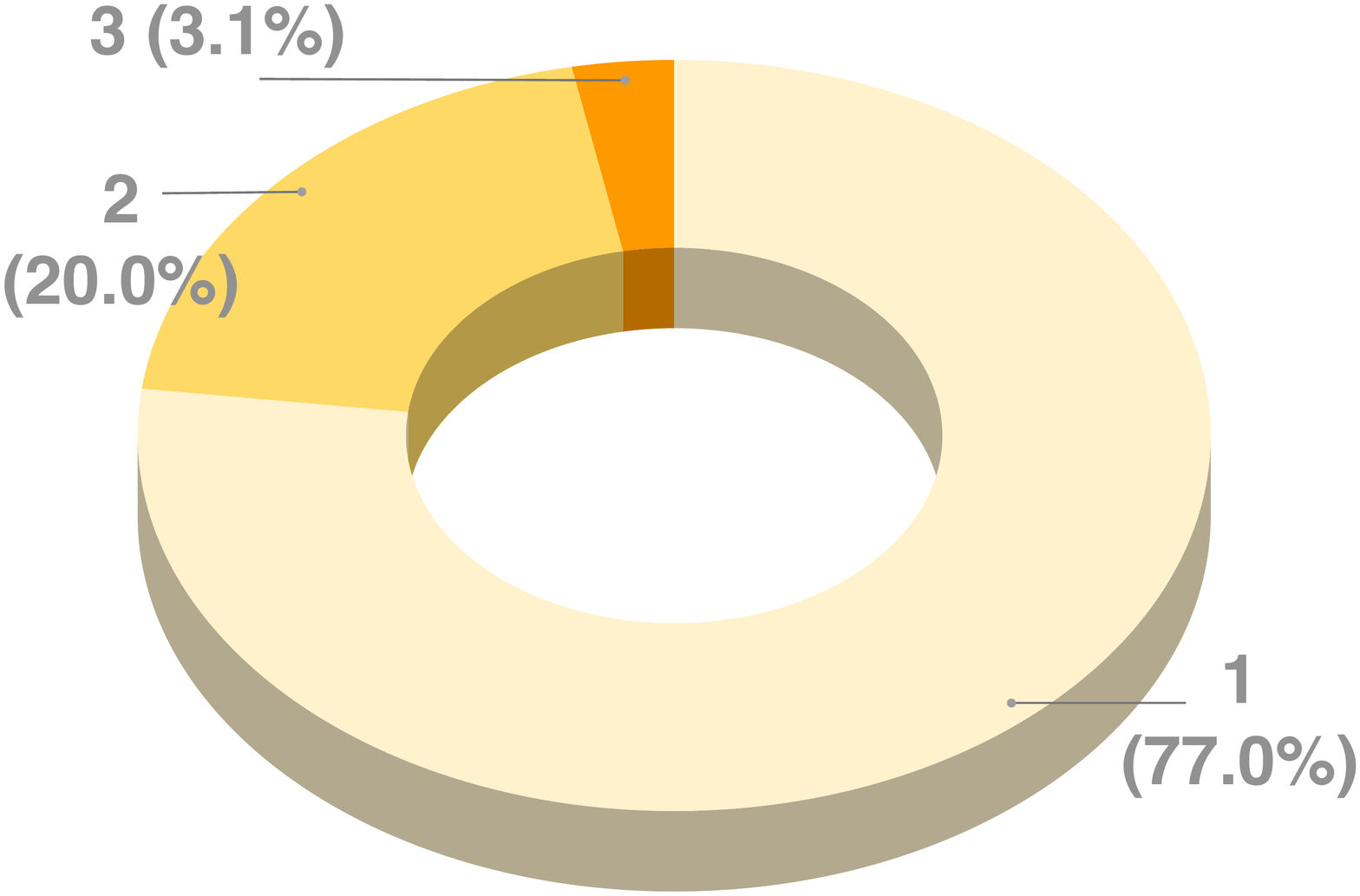}
        \caption{\centering \#words per keyphrase}
    \end{subfigure}
  \caption{Frequency and length distribution of keyphrases.}
  \label{fig:keyword_dist}
  \vspace{-10pt}
\end{figure}

\begin{figure}[ht]
\centering
  \includegraphics[width=0.46\textwidth]{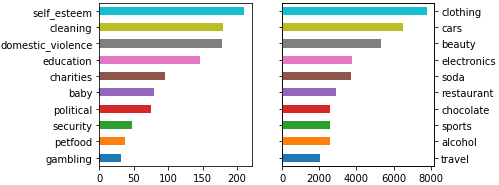}
  \caption{Distribution of images per category (top $10$ categories by count on the right, bottom $10$ on left).}
  \label{fig:cat_image_dist}
  \vspace{-10pt}
\end{figure}

\begin{figure}[ht]
\centering
  \includegraphics[width=0.46\textwidth]{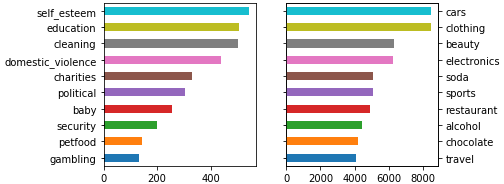}
  \caption{Distribution of unique phrases per category (top $10$ categories by count on the right, bottom $10$ on left).}
  \label{fig:cat_uphrase_dist}
  \vspace{-20pt}
\end{figure}

The minimum, mean and maximum number of images associated with a brand are 1, 19 and 282 respectively. The top three categories of advertisements are \emph{clothing}, \emph{cars} and \emph{beauty products} with 7798, 6496 and 5317 images respectively. Least number of advertisements are associated with \emph{gambling} (32), \emph{pet food} (37) and \emph{security and safety services} (47) respectively.
Additional statistics around the dataset (\emph{i.e.}, keyphrase lengths, images per category, and unique keyphrases per category) are shown in Figures~\ref{fig:keyword_dist}, \ref{fig:cat_image_dist}, and \ref{fig:cat_uphrase_dist}.

\subsection{Evaluation Metrics}
We use different evaluation metrics to measure performance of our classification and ranking models. We use three different metrics to evaluate the performance of each model.

\subsubsection{Classification metrics}
We rely on accuracy, text similarity and set-intersection based recall to evaluate model performance. 
\begin{itemize}
\item \textbf{Accuracy.} We predict the keyphrase with the highest probability for each image and match it with the labels (ground truth keyphrases) for the image. We use the score of the matched phrase as the accuracy. If no labels match for a sample, the accuracy is $0$.
We average the accuracy scores over all the test instances to report test accuracy.

\item \textbf{Similarity:} Accuracy neglects the semantic similarity between the predicted phrase and the labels. For example, a predicted keyphrase ``a great offer'' is similar to one of the labels, ``great sale'', but will gain $0$ for accuracy.
So we calculate the cosine similarity~\cite{han2011data} between the embeddings of the predicted keyphrase and each label. Then we multiply the similarity scores with each label's score and keep the maximum as the final similarity score for the sample.

\item \textbf{VQA Recall$@3$:} we use Recall at $3$ ($R_{VQA}$@$3$) as an evaluation metric for the classification task (essentially like the VQA formulation in \cite{kovashka_eccv2018}). For each test instance $i$, the ground truth is limited to top $3$ keyphrases 
leading to set $\mathcal{K}^*_i$. From the classification model's predictions the top $3$ keyphrases are chosen leading to set $\hat{\mathcal{K}}_i$. $R_{VQA}@3$ is simply $\frac{|\hat{\mathcal{K}}_i \cap \mathcal{K}^*_i |}{3}$.
\end{itemize}

\subsubsection{Ranking metrics} 
We use the same evaluation metrics from prior work  \cite{self_recsys2019}, mainly precision (P@K), recall (R@K), and NDCG$@$K \cite{ndcg2002} to evaluate the proposed ranking model. 
It is worth noting that recall is computed differently for evaluating ranking and classification models proposed in this work.  
Formally, given a set of queries $\mathcal{Q} = \{q_1 \cdots q_n\}$, set of phrases $\mathcal{D}_i$ labeled \emph{relevant} for each query $q_i$ and the set of relevant phrases $\mathcal{D}_{ik}$ retrieved by the model for $q_i$ at position $k$, we define $R@K = \frac{1}{N} \cdot \sum_{i=1}^{N}\frac{|\mathcal{D}_{ik}|}{|\mathcal{D}_i|}  $.

\subsection{Implementation details}
For the classification model, we set the number of object-relationship, language, and cross-modality layers as $5$, $9$, $5$, and leverage pre-trained parameters from \cite{lxmert}. We fine-tune the encoders with our dataset for $4$ epochs. The learning rate is $5e^{-5}$ (adam optimizer), and the batch size is $32$. We also set $D_t, D_p$, and $D_e$ equal to $768$.
For the similarity evaluation, we average the GloVe~\cite{glove} embeddings of all the words in a phrase to calculate the phrase embedding. For the DRMM model (ranking formulation), we used the MatchZoo implementation \cite{matchzoo}, with $300$ for batch size, $10$ training epochs, $10$ as the last layer's size, and learning rate of $1.0$ (adadelta optimizer).
We combine textual data from different sources in a consistent order separated by a [SEP] symbol before feeding them into the encoder.

\begin{figure*}[ht]
\centering
  \includegraphics[width=0.6\textwidth]{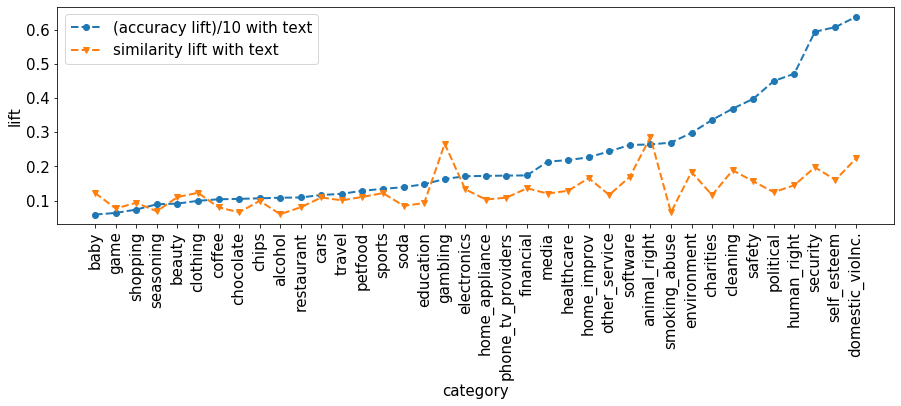}
  \caption{Performance lifts across different categories after using text features. For accuracy, the lift is scaled (divided by $10$) for better visualization with the similarity lift.}
  \label{fig:category_perf}
\end{figure*}

\begin{table}[h]
    \centering
    \begin{tabular}{|l|c|c|c|}
    \hline 
     features & accuracy (\%) & similarity (\%) & $R_{VQA}$@$3$ \\
     \hline 
     \hline 
    I  & 10.05 & 58.05 & 0.447 \\ 
    I$\times$Q & 12.18 & 58.26 & 0.450 \\ 
    I$\times$(Q+W) & 19.01 & 60.12 & 0.467 \\ 
    I$\times$(Q+O) & 19.50 & 60.34 & 0.470 \\ 
    I$\times$(Q+W+O) & \textbf{20.40} & \textbf{60.95} & \textbf{0.473} \\ 
    \hline
    Q+W+O & 13.39 & 60.13 & 0.450 \\
    \hline
    non cross-modal & 18.65  & 60.68 & 0.460 \\ 
    \hline
    \end{tabular}
    \caption{Classification performance with different features (I: image, Q: question, W: Wikipedia page of associated brand, and O: OCR, \emph{i.e.}, text in ad image, $\times$: cross-modal representation); non cross-modal denotes using an addition of separate visual and linguistic features.}
    \label{tab:classification_results}
    \vspace{-20 pt}
\end{table}

\vspace{-10 pt}
\subsection{Results} 
For different sets of multimodal features, the performance results are reported in Table~\ref{tab:classification_results} (for classification) and Table~\ref{tab:ranking_results} (for ranking) respectively.
The presence of Wikipedia and OCR text gives a significant lift over using only the image. Both classification and ranking metrics show the same trend in terms of feature sets. 

Table~\ref{tab:classification_results} shows that linguistic features dramatically lift the performance by $103\%$, $5\%$, and $6\%$ in accuracy, similarity, and $R_{VQA}$@$3$, compared to the performance of the model trained \emph{only} with visual features while only using linguistic features (Q+W+O) causes a big drop in all the performances. It occurs that the OCR features bring more performance lift, compared to the Wiki. We think  knowing more about the brand with the Wikipedia pages is beneficial to recommend themes to designers~\cite{self_recsys2019} while the written texts on the images (OCR) are sometimes more straightforward for recommendations.  
In addition, as reported in Table~\ref{tab:classification_results} (non cross-modal), using separate text and image embeddings (obtained from the model in Figure~\ref{fig:lxmert_model}) is inferior in performance compared to the cross-modal embeddings.
We notice that the accuracy scores are comparatively low; this reflects the difficult nature of understanding visual ads~\cite{cvpr_kovashka}. 

In Table~\ref{tab:ranking_results}, we observe very similar patterns: OCR features bring more benefits to ranking than Wikipedia pages. We notice that in P@$10$ and R@$5$, only using image feature (I) achieves a better score compared to adding the question features. This may indicate that the local interactions in DRMM are not effective with short questions, but favor longer textual inputs such as OCR and Wikipedia pages.

\begin{table}[h]
    \centering
    \begin{tabular}{|l|c|c|c|c|c|c|}
    \hline 
    \multirow{2}{*}{Features}  & \multicolumn{2}{|c|}{Precision} & \multicolumn{2}{|c|}{Recall}  & 
    \multicolumn{2}{|c|}{NDCG}  \\ \cline{2-7}
      & @$5$  & @$10$   &  @$5$ & @$10$ & @$5$ & @$10$   \\
     \hline 
     \hline 
      I &   0.150 &  0.126 & 0.161 &  0.248 & 0.158 & 0.217 \\
      I$\times$Q & 0.152  & 0.124  & 0.158 &  0.259 & 0.162 & 0.227  \\
      I$\times$(Q+W) & 0.154  & 0.130  &   0.160 &  0.271 & 0.161 &  0.234  \\
      I$\times$(Q+O) & 0.174  &  0.137 &   0.182 &   0.287 & 0.185 & 0.254  \\
      I$\times$(Q+W+O) & \textbf{0.183} & \textbf{0.141} & \textbf{0.191} & \textbf{0.294} & \textbf{0.198} & \textbf{0.265} \\
    \hline 
    \end{tabular}
    \caption{Ranking performance with different features (I: image, Q: question, W: text from brand Wikipedia page, and O: OCR text in the ad image, $\times$: cross-modal representation).}
    \label{tab:ranking_results}
    \vspace{-20 pt}
\end{table}

\subsection{Insights}
\begin{figure}[ht]
\centering
  \includegraphics[width=0.4\textwidth]{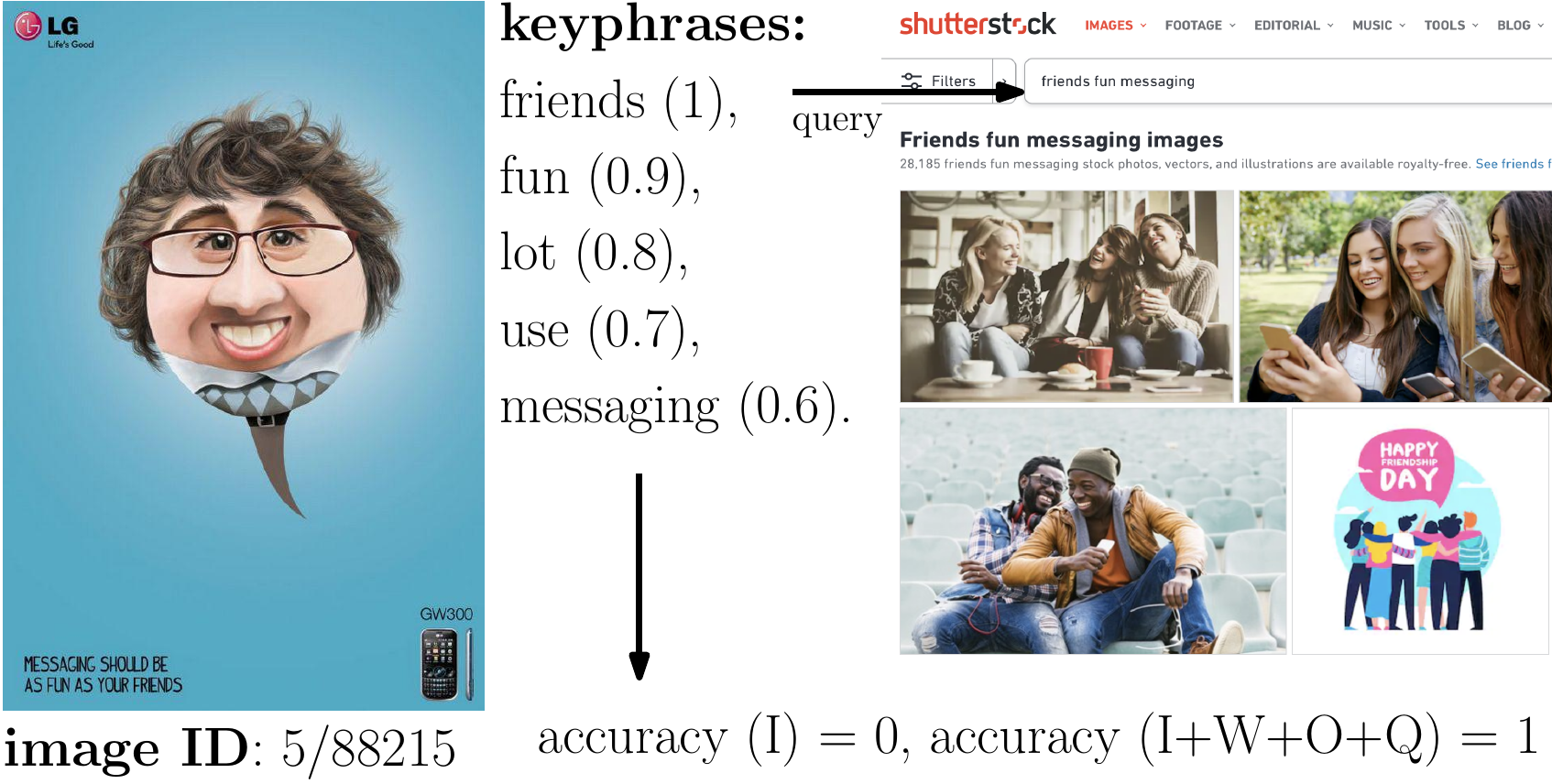}
  \caption{The ad image on the left is a sample in the public dataset \cite{creative_dataset}, and the ground truth keyphrases with scores are as shown. In the classification setup, using only the image has zero accuracy, while using image + text features leads to perfect accuracy. The predicted keyphrases can be used as recommended queries to a stock image library \cite{shutterstock} (as shown on the right) to obtain new creatives.}
  \label{fig:insights_example}
\end{figure}

Figure~\ref{fig:category_perf} shows the performance lifts in accuracy and similarity metrics per category (where lift is defined as ratio of improvement to baseline result without using text features in the classification task).
As shown, multiple categories, \emph{e.g.}, public service announcement (PSA) ads around domestic violence and animal rights benefit from the presence of text features; this may be related to the hidden symbolism \cite{kovashka_eccv2018} common in PSAs, where the text can help clarify the context even for humans. Also, similarity and accuracy metrics do not have the same trend in general.
Along the lines of inferring themes from past ad campaigns, and assisting strategists towards designing new creatives, we show an example based on our classification model in Figure~\ref{fig:insights_example}. In general, a strategist can aggregate recommended keyphrases across a brand or product category, and use them to design new creatives.

\vspace{-12 pt}
\section{Conclusion} \label{sec:conclusion}
In this paper, we make progress towards automating the inference of themes (keyphrases) from past ad campaigns using multimodal information (\emph{i.e.}, images and text). In terms of model accuracy, there is room for improvement, and using generative models for keyphrase generation may be a promising direction. In terms of application, \emph{i.e.},
automating creative design, we believe that the following are natural directions for future work: (i) automatically selecting ad images and generating ad text based on recommended themes, and (ii) launching ad campaigns with new creatives (designed via our proposed method) and learning from their performance in terms of CTR and CVR. Nevertheless, the proposed method in this paper can increase diversity in ad campaigns (and potentially reduce ad fatigue), reduce end-to-end design time, and enable faster exploratory learnings from online ad campaigns by providing multiple themes per brand (and multiple images per theme via stock image libraries).

\bibliographystyle{abbrv}
\bibliography{refs}

\end{document}